\documentclass[10pt,twocolumn,letterpaper]{article}

\usepackage[pagenumbers]{cvpr} 

\usepackage[dvipsnames]{xcolor}

\definecolor{cvprblue}{rgb}{0.21,0.49,0.74}
\usepackage[pagebackref,breaklinks,colorlinks,citecolor=cvprblue]{hyperref}
\usepackage{multirow}
\usepackage{csquotes}
\newcommand{\ours}{{\sc InstantSwap}\xspace}
\newcommand{\csb}{{\textit{ConSwapBench}}\xspace}

\usepackage{color}
\definecolor{color3}{rgb}{0.95,0.95,0.95}
\definecolor{Red}{RGB}{192, 0, 0}
\definecolor{Blue}{RGB}{12, 114, 186}
\definecolor{Yellow}{RGB}{218, 169, 20}

\def\paperID{*****} 
\def\confName{CVPR}
\def\confYear{2024}

\title{\ours: Fast Customized Concept Swapping across Sharp Shape Differences}

\author{Chenyang Zhu$^{1, *}$\qquad\quad Kai Li$^{2, *, \dagger}$\qquad\quad Yue Ma$^{3, *}$ \qquad\quad Longxiang Tang$^1$ \\ Chengyu Fang$^1$\qquad\quad Chubin Chen$^1$ \qquad\quad  Qifeng Chen$^{3}$\qquad\quad Xiu Li$^{1, \dagger}$\\
$^1$ Tsinghua University \qquad $^2$ Meta Reality Labs \qquad $^3$ HKUST\\
\url{https://instantswap.github.io/}
}

\begin{document}
\twocolumn[{
\maketitle

\begin{center}
\captionsetup{type=figure}
\includegraphics[width=\textwidth]{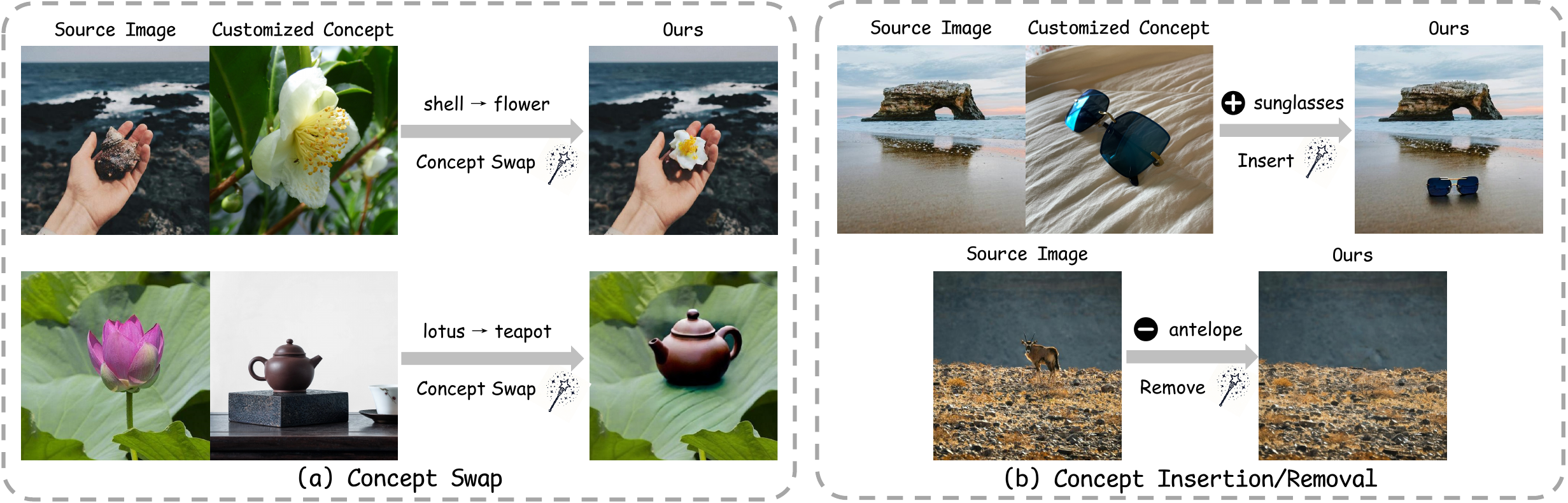}
\captionof{figure}{
Visual results of \ours. Our approach can seamlessly swap a source concept with a customized concept in an image, even with great shape differences. Moreover, \ours can be used for other tasks, such as concept insertion and removal.
}
\label{fig:title case}
\end{center}
}]

\renewcommand{\thefootnote}{\fnsymbol{footnote}}
\footnotetext{$*$~Equal contribution.}
\footnotetext{$\dagger$~Corresponding Authors.}

\begin{abstract}
Recent advances in Customized Concept Swapping (CCS) enable a text-to-image model to swap a concept in the source image with a customized target concept.
However, the existing methods still face the challenges of \textit{\textbf{inconsistency}} and \textit{\textbf{inefficiency}}. They struggle to maintain consistency in both the foreground and background during concept swapping, especially when the shape difference is large between objects. 
Additionally, they either require time-consuming training processes or involve redundant calculations during inference.
To tackle these issues, we introduce \ours, a new CCS method that aims to handle sharp shape disparity at speed.
Specifically, we first extract the bbox of the object in the source image \textit{automatically} based on attention map analysis and leverage the bbox to achieve both foreground and background consistency. For background consistency, we remove the gradient outside the bbox during the swapping process so that the background is free from being modified. 
For foreground consistency, we employ a cross-attention mechanism to inject semantic information into both source and target concepts inside the box. 
This helps learn semantic-enhanced representations that encourage the swapping process to focus on the foreground objects.
To improve swapping speed, we avoid computing gradients at each timestep but instead calculate them periodically to reduce the number of forward passes, which improves efficiency a lot with a little sacrifice on performance. 
Finally, we establish a benchmark dataset to facilitate comprehensive evaluation. Extensive evaluations demonstrate the superiority and versatility of \ours.
\end{abstract}    
\section{Introduction}
\label{sec:intro}
\begin{figure*}[tb]
  \centering
  \includegraphics[width=\linewidth]{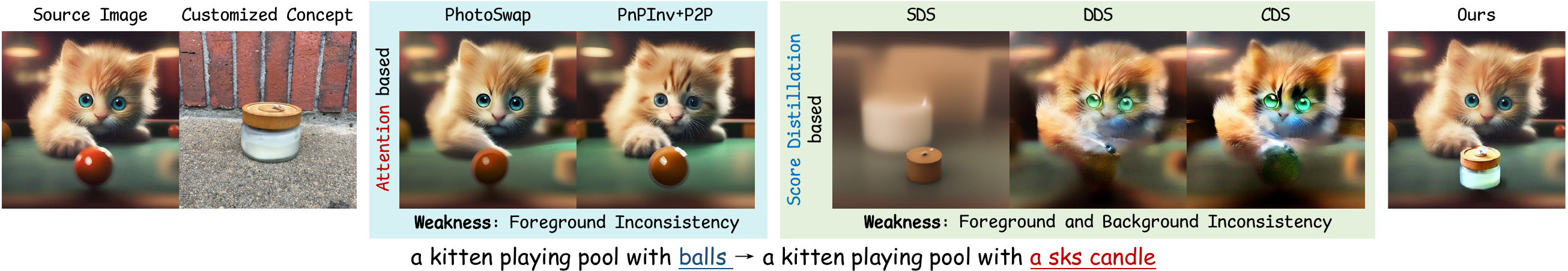}
  \caption{Our \ours achieves better swapping consistency than the existing methods.} 
  \label{fig:intro case}
\end{figure*}
We explore the task of Customized Concept Swapping (CCS), a subtask of text-to-image (T2I) generation, which aims to replace a concept in a source image with a highly customized new concept.
Combined with diffusion models~\citep{diffbgan,SD,glide}, recent CCS methods demonstrate widespread applicability in areas such as selfie enhancement, photo blog creation, and comic creation.

Early work~\citep{paint} in CCS primarily relies on copy-paste techniques, which are rough and unreliable.
By integrating powerful customization techniques~\citep{db,cd} with image editing methods~\citep{p2p,dds}, a series of works~\citep{photoswap,customedit,dreamedit,swapanything} have been proposed.
Although achieving remarkable success, these approaches still face the problems of \textit{inconsistency} and \textit{inefficiency} as shown in \cref{fig:intro case}.
(1) \textit{Inconsistency}: Attention-based methods such as PhotoSwap~\citep{photoswap} and P2P~\citep{p2p} maintain background consistency well but struggle with shape differences between source and target concepts, resulting in foreground inconsistency. 
Score distillation based methods such as SDS~\citep{sds}, DDS~\citep{dds}, and CDS~\citep{cds} fail to generate foreground concepts precisely and alter the background significantly, causing both foreground and background inconsistency. 
(2) \textit{Inefficiency}: Attention-based methods require an inefficient training phase~\citep{null,pnpinversion} on source image to maintain background consistency.
While score distillation based methods are training-free, they still require redundant calculations of forward passes at each timestep, leading to inference inefficiency.

To address the aforementioned issues, we propose \ours, a training-free framework that \textit{\textbf{efficiently}} performs customized concept swapping across shape differences while maintaining \textit{\textbf{both foreground and background consistency}}. 
Specifically, we extract the bounding box (bbox) that indicates the position of the source concept from the enhanced cross-attention map of the source image.
With this bbox, we perform the background gradient masking (BGM) strategy to prevent modifications outside the bbox, thus ensuring background consistency. 
Moreover, to improve foreground consistency, we leverage the semantic information to highlight the cross-attention maps of source and target concepts respectively within the bbox. This strategy leads to the semantic-enhanced concept representation (SECR), which facilitates precise foreground swapping.  
Finally, we introduce the Step-skipping Gradient Updating (SSGU) strategy, which only performs forward passes at certain timesteps to calculate gradients. 
For the timesteps without direct gradient computations, we reuse the previously obtained gradients for updates.
Through this strategy, we reduce the total number of forward passes and improve the efficiency of our method.

Since the CCS is a recently proposed task, no dedicated evaluation benchmark currently exists. To address this gap, we introduce \csb, the first benchmark dataset specifically designed for CCS. \csb comprises two sub-benchmarks: ConceptBench and SwapBench. ConceptBench contains images representing target concepts, while SwapBench includes images with one or more concepts to be swapped, serving as source images.

Through extensive qualitative and quantitative comparisons, we demonstrate the effectiveness and superiority of our \ours. We also conduct comprehensive ablation studies to verify the effectiveness of each component of our approach. Additionally, we further extend our \ours to related tasks, proving its efficacy and versatility.
Our contributions are summarized as follows:

\begin{itemize}
    \item We propose \ours, a novel training-free customized concept swapping (CCS) framework, which enables efficient concept swapping across sharp shape differences.
    
    \item We design the background gradient masking (BGM) strategy and semantic-enhanced concept representation (SECR) to improve the background and foreground consistency respectively. Moreover, we adopt a step-skipping gradient updating (SSGU) strategy to reduce redundant computation and improve efficiency.

    \item To provide a comprehensive evaluation for CCS, we introduce \csb, the first benchmark for customized concept swapping. Extensive qualitative and quantitative evaluations demonstrate the effectiveness and superiority of our \ours.
\end{itemize}

\begin{figure*}[tb]
  \centering
  \includegraphics[width=\linewidth]{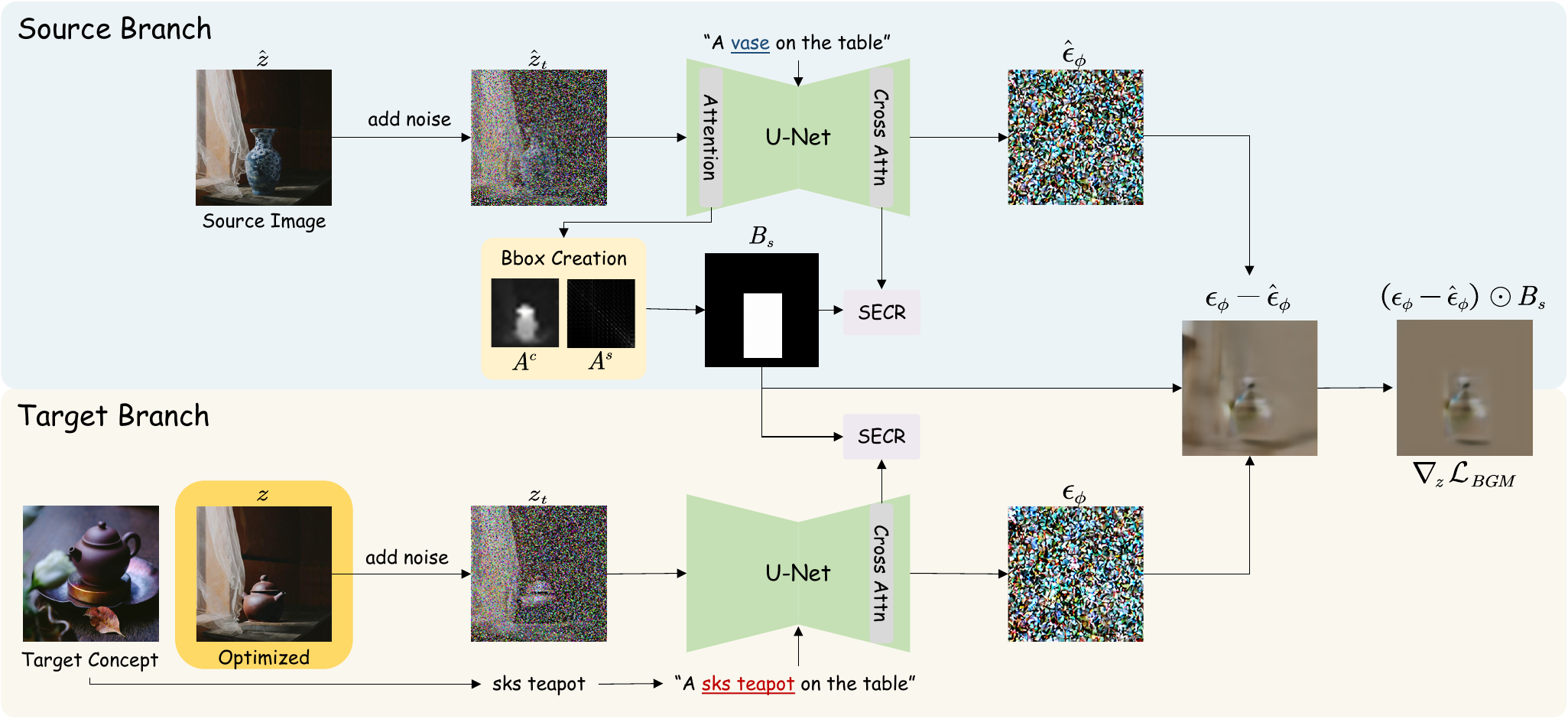}
  \caption{Overall pipeline of \ours. We first obtain the bbox of the source concept automatically. The obtained bbox is input into SECR in both the source and target branches to enhance the foreground swapping consistency. Additionally, the source and target branches generate the prediction of noise for the source and target images based on their respective prompts. The predicted noise, along with the bbox, is used for the BGM to preserve background consistency.} 
  \label{fig:frame}
\end{figure*}

\section{Related work}

\subsection{Diffusion-based Image Editing}

Image Editing is a fundamental and popular topic in computer vision. Previous works based on Generative Adversarial Networks (GAN)~\citep{gan} only focus on specific object domains, which limits the application. With the emergence of diffusion model~\cite{SD}, image editing is now able to modify various objects through prompts. 
These methods are mainly divided into five categories: instruction-based methods, blending-based, attention-based, inversion-based, and score distillation based methods.
Instruction-based methods~\citep{instructpix2pix,instructdiffusion,smartedit,magicbrush, ma2024followyourpose} typically require an instruction editing dataset to train the diffusion model. 
Blending-based methods~\citep{diffedit,tuning,forgedit,huang2023pfb} merge the source and target prompts to guide the editing process, while attention-based methods~\citep{p2p,masactrl,pnp,guo2024focus} inject the attention feature of the source image. 
Both methods have lower editing costs but poorer background preservation and prompt alignment. 
Inversion-based methods~\citep{null,pnpinversion,negative,dong2023prompt, ma2024followyourclick} aim to reverse the fixed trajectory generated by the forward pass to reproduce the source image. These methods can serve as an extra training phase to enhance the background consistency of attention-based methods.
Finally, score distillation based methods~\citep{dds,cds,groundscore,collaborative, ma2023magicstick, ma2024followyouremoji} draw on the optimization process of SDS~\citep{sds}, using score distillation-based loss to optimize the source image for editing. 
These methods are more flexible than the previous ones but still face challenges with background preservation.

\subsection{Concept Swapping}

Concept swapping, a subtask of general image editing, focuses on replacing the source concept in an image with a user-specified target concept. 
This task is first proposed by PbE~\citep{paint}, which employs a CLIP encoder to extract features of the target concept and inject them into the UNet through a cross-attention layer. 
After that, concurrent works~\citep{photoswap,customedit,wang2024cove, dreamedit} extend concept swapping into the customization field~\citep{TI,multibooth}.
They combine attention-based editing methods~\citep{p2p,pnp}, with tuning-based customization methods~\citep{db,cd} to achieve customized concept swapping.
Building on Photoswap, SwapAnything~\citep{swapanything} further obtains masks with external modules to specify the locations of objects in the source image. 
We improved the existing method in three key aspects.
Firstly, we employ bounding boxes instead of masks as spatial indicators of the source concept, allowing greater flexibility for shape variation during concept swapping.
Secondly, we use bounding boxes to prevent background changes via gradient masking, thus ensuring background consistency. Third, we augment concept representation with semantic information to maintain foreground consistency.
Finally, rather than executing forward passes at every timestep, we execute them only at specific intervals to enhance efficiency.
\section{Method}
Given a set of images, $\mathcal{X}_t=\{ x_i \}_{i=1}^M$ representing a specific concept $O_t$, along with an image $x_s$ and a prompt $p_{s}$ describing a source concept $O_s$, the objective of CCS is to ``seamlessly'' replace $O_s$ in $x_s$ with $O_t$ according to a target prompt $P_{t}$, resulting in a final target image $x_t$. 
An ideal customized concept swapping should handle the shape differences between source and target concepts to preserve swapping consistency while maintaining satisfactory efficiency. We introduce \ours to achieve this. \ours is based on Stable Diffusion and extends from the score distillation based image editing methods \citep{sds,dds}.

\subsection{Preliminaries}

\begin{figure*}[t]
  \centering
  \includegraphics[width=0.9\linewidth]{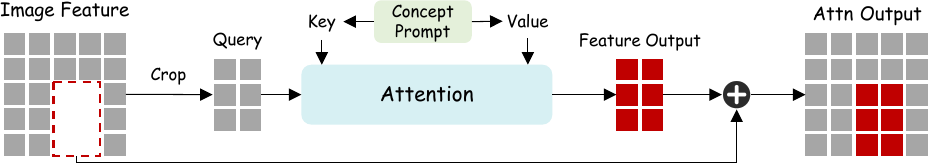}
  \caption{Overview of SECR. 
  }
  \label{fig:SECR overview}
\end{figure*}

\subsubsection{Stable Diffusion}

In this paper, the foundational model utilized for text-to-image generation is Stable Diffusion~\citep{SD}. 
It takes a text prompt $P$ as input and generates the corresponding image $x$. Stable Diffusion consists of three main components: an autoencoder$(\mathcal{E}(\cdot),\mathcal{D}(\cdot))$, a CLIP text encoder $\tau(\cdot)$ and a U-Net $\epsilon_{\phi}(\cdot)$.
Typically, it is trained with the guidance of the following reconstruction loss:
\begin{equation}
\label{loss:rec}
\mathcal{L}_{rec}=\mathbb{E}_{z,\epsilon \sim \mathcal{N}\left( 0,1 \right) ,t,P}\left[ \lVert \epsilon -\epsilon _{\phi}\left( z_t,t,\tau\left( P \right) \right) \rVert_2^2 \right],
\end{equation}
where $\epsilon \sim \mathcal{N}\left( 0,1 \right)$ is a randomly sampled noise, t denotes the time step. 
The calculation of $z_t$ is given by $z_t=\alpha_t z+\sigma_t \epsilon$, where the coefficients $\alpha_t$ and $\sigma_t$ are provided by the noise scheduler. 

\subsubsection{Score Distillation Based Image Editing}

Different from traditional attention-based image editing, score distillation based methods achieve image editing through iterative optimization with a score distillation loss.
Given the latent feature $z$ of source image and a denoising U-Net $\epsilon_{\phi}(\cdot)$, SDS~\citep{sds} can optimize the latent feature $z$ of the image to align with the target prompt $P_{t}$ by employing the following loss:
\begin{equation}
    \mathcal{L}_{SDS}=\lVert \epsilon _{\phi}\left( z_t,t,\tau\left( P_{t} \right) \right)- \epsilon \rVert_2^2,
\label{equ:sds loss}
\end{equation}
where $\epsilon$ and $t$ are randomly sampled noise and timestep.

The resulting image SDS is very blurry and only contains foreground objects in the target prompt $P_{t}$. 
To address this issue, DDS~\citep{dds} expresses the gradient of \cref{equ:sds loss} as
\begin{equation}
\label{equ:sds target}
    \nabla_z\mathcal{L}_{{SDS}}(z_t, t, \tau(P_{t}))=\delta_{ {tgt }}+\delta_{ {bias }},
\end{equation}
where $\delta_{ {tgt}}$ indicates the direction aligned with the target prompt and $\delta_{ {bias }}$ refers to undesired part that makes the image blurry.
Based on this, DDS further utilizes the fixed latent $\hat{z_t}$ of the source image and the source prompt $P_{s}$ to approximate the bias component in \cref{equ:sds target}:
\begin{equation}
\label{equ:sds source}
    \nabla_z\mathcal{L}_{{SDS}}(\hat{z_t}, t, \tau(P_{s})) \approx \hat{\delta}_{ {bias }}\approx \delta_{ {bias }}.
\end{equation}
Finally, DDS is represented by the difference of \cref{equ:sds target} and \cref{equ:sds source}:
\begin{align}
\label{equ:sds approx}
\nabla_z\mathcal{L}_{{DDS}} &= \nabla_z\mathcal{L}_{{SDS}}(z_t, t, \tau(P_{t})) \notag \\
&\quad - \nabla_z\mathcal{L}_{{SDS}}(\hat{z_t}, t, \tau(P_{s})) \notag \\
&\approx \delta_{ {tgt }}.
\end{align}

Based on \cref{equ:sds approx}, the loss of DDS is given by
\begin{equation}
\label{equ:dds}
    \mathcal{L}_{DDS}=\lVert \epsilon _{\phi}\left( z_t,t,\tau\left( P_{t} \right) \right)- \hat{\epsilon} _{\phi}( \hat{z_t},t,\tau( P_{s} ) ) \rVert_2^2.
\end{equation}

\subsection{\ours}

Directly extending score distillation based editing methods to the task of CCS encounters the challenge of inconsistency.
These methods optimize the background and foreground simultaneously, causing cross-interference and leading to undesirable inconsistency.  
To address these limitations, we first propose a strategy to \textit{automatically} locate objects to be edited, resulting in the object bounding box (bbox). With this bbox, we propose a background gradient masking technique to remove gradients in the background region and confine swapping to the foreground region. 
To further enhance foreground swapping consistency, we propose to learn Semantic-enhanced concept representations for both source and target concepts based on an attention map feature injection mechanism. An overview of our method is presented in \cref{fig:frame}.

\subsubsection{Automatic Bounding Box Generation}\label{sec:bbox}
We first automatically obtain the bbox to indicate the position of the concept $O_s$ in the source image.
Given the source image $x_s$ and the source prompt $P_{s}$, we perform a forward pass with the U-Net $\epsilon_{\phi}(\cdot)$ and obtain the cross-attention map $A^c$ and self-attention map $A^s$ through:
\begin{equation}
A=\operatorname{Softmax}\left(\frac{Q K^T}{\sqrt{d^{\prime}}}\right) V, 
\label{equ:attn map}
\end{equation}
where $Q$ is the query vector projected from the image features, $d^\prime$ represents the output dimension of key and query features.
$K$ is the key vector and $V$ is the value vector.
For cross-attention maps $A^c$, $K$ and $V$ are projected from the text embeddings $\tau(P)$.
For self-attention maps $A^s$, $K$, and $V$ are projected from the image features.
Directly applying a threshold on the $A^c$ can yield a coarse-grained mask, which cannot accurately reflect the location of $O_s$. Inspired by~\citep{datasetdiff,attn-calib}, we modify the $A^c$ as follows:
\begin{equation}
    \hat{A}^c=A^s\cdot \left( A^c \right) ^{\alpha}.
\end{equation}
Based on \cref{equ:attn map}, all values in $A^c$ range between 0 and 1. 
Therefore, element-wise exponentiation of $A^c$ by $\alpha$ can weaken the activation of non-target regions. Additionally, as mentioned in~\citep{towards}, $A^s$ contains rich structural information.
This information can effectively assist $\hat{A}^c$ in better activating the target regions.
Finally, we apply the threshold $\beta$ to $\hat{A}^c$ to obtain the mask.
Subsequently, we converted the mask into the bbox $B_s$ based on the minimum and maximum coordinates of all foreground points within the mask.
This strategy allows us to obtain the bbox $B_s$ without any additional modules.
We intentionally set a relatively loose constraint on the mask to obtain a bbox that completely covers the source concept.
We discuss the effectiveness of our automatically obtained bboxes in \cref{sec:ablation bbox}.

\subsubsection{Background Gradient Masking}\label{sec:BGM}

With the object bbox, we propose a background gradient masking (BGM) approach to ensure that the concept swap is confined to the foreground region. Given the latent feature $\hat{z}$ of the source image and the latent feature $z$ of the target image, where $z$ is initialized to $\hat{z}$ and is continuously optimized to obtain the final target image $x_t$.
Based on \cref{equ:dds}, we first obtain the gradient of $z$:
\begin{equation}
    \nabla_{z} \mathcal{L}_{}=(\epsilon _{\phi}\left( z_t,t,\tau\left( P_{t} \right) \right)- \hat{\epsilon} _{\phi}( \hat{z_t},t,\tau( P_{s} ) )\frac{\partial \epsilon _{\phi}\left( z_t,t,\tau\left( P_{t} \right) \right)}{\partial z_t}\frac{\partial z_t}{\partial z}.
\end{equation}
As stated in~\citep{sds}, the mid term is a U-Net Jacobian term and can be omitted, and $\alpha_t=\partial z_t \ / \partial z$ is a constant which can be represented as $w(t)$:
\begin{equation}
    \nabla_{z} \mathcal{L}_{}=w(t)(\epsilon _{\phi}\left( z_t,t,\tau\left( P_{t} \right) \right)- \hat{\epsilon} _{\phi}( \hat{z_t},t,\tau( P_{s} ) ).
    \label{equ:dds gradient}
\end{equation}
This gradient shares the same dimension as $z$, which means it can update $z$ in a pixel-wise manner.
However, this will update the foreground and background simultaneously, producing inconsistent background. 
To remedy this, we apply the bbox $B_s$ on 
\cref{equ:dds gradient} to mask the gradients related to the background before back propagation and obtain our BGM:
\begin{equation}
    \nabla_{z} \mathcal{L}_{BGM}=w(t)(\epsilon _{\phi}\left( z_t,t,\tau\left( P_{t} \right) \right)- \hat{\epsilon} _{\phi}( \hat{z_t},t,\tau( P_{s} ) )\odot B_s.
    \label{equ:BGM}
\end{equation}
This simple masking strategy prevents the background from being updated and thus ensures background consistency.

\begin{figure*}[ht]
  \centering
  \includegraphics[width=\linewidth]{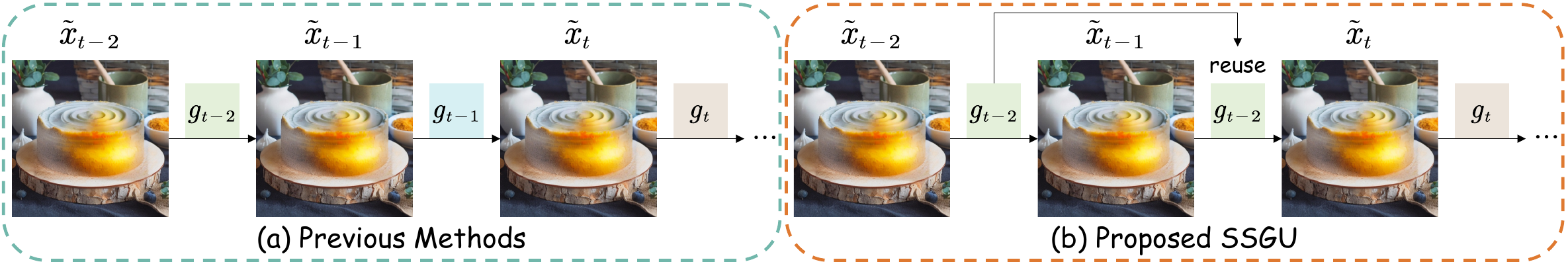}
  \caption{Comparison between our SSGU and previous methods. 
  }
  \label{fig:SSGU}
\end{figure*}

\subsubsection{Semantic-enhanced Concept Representation}\label{sec:SECR}

The BGM module maintains the background consistency during swapping. 
However, whether the source concept can be replaced with the target concept cannot be guaranteed. 
This limitation arises because the optimization of \cref{equ:BGM} is still carried out at the entire feature map level of both the source latent $\hat{z_t}$ and target latent $z_t$ without distinguishing between the foreground and the background. 
To address this, we propose to obtain Semantic-enhanced concept representations for both source and target concepts and emphasize their locations within the foreground region during concept swapping.

Let $F_s$ be the source image feature and $p_s$ represent the prompt of source concept (e.g. \enquote{rose}), the semantic embedding $c_s$ can be acquired through $c_s=\tau (p_s)$.
We first resize the previously obtained object bbox to fit the dimensions of the source image feature $F_s$, resulting in the feature bbox $B_f$.
We then crop $F_s$ with $B_f$ to get a regional image feature ${f}_s$.
With ${f}_s$, we calculate the query vector through $Q_s=W^q\cdot {f}_s$.
After that, we can obtain the key and value vectors through:
\begin{equation}
    K_s=W^k\cdot c_s, V_s=W^v\cdot c_s.
\end{equation}
Then the final partial attention output is calculated as follows:
\begin{equation}
\hat{f_s}=\operatorname{Softmax}\left(\frac{Q_s K_s^T}{\sqrt{d^{\prime}}}\right) V_s,
\end{equation}
where $d^\prime$ represents the output dimension of key and query features. 
In this way, we inject the semantic information of the source concept into the cross-attention map, resulting in regional concept representation $\hat{f_s}$.
We then map $\hat{f_s}$ back to the original feature map $F_s$ to get a Semantic-enhanced representation $\hat{F}_s$ for the entire source image. \cref{fig:SECR overview} illustrates the process. 
In the target branch, we first convert the target concept into semantic space with DreamBooth, using a specific rare token (e.g., \enquote{\texttt{sks}}) to represent the concept.
With the target prompt $p_t$ (e.g., \enquote{\texttt{sks teapot}}) and the feature bbox $B_f$, we similarly apply this process for the target image feature $F_t$ and obtain the semantic-enhanced representation $\hat{F}_t$ for the target image. 

Through proactive injection of semantic guidance, we provide the source and target branches with Semantic-enhanced concept representation within the foreground region.
Consequently, SECR transforms the target branch into a target concept adder and the source branch into a source concept remover.
Their collaboration results in precise and seamless concept swapping, thus enhancing the foreground consistency. 
Moreover, SECR can also facilitate concept insertion and removal, which is further discussed in \cref{sec:concept insert remove}.

\begin{figure*}[tb]
  \centering
  \includegraphics[width=\linewidth]{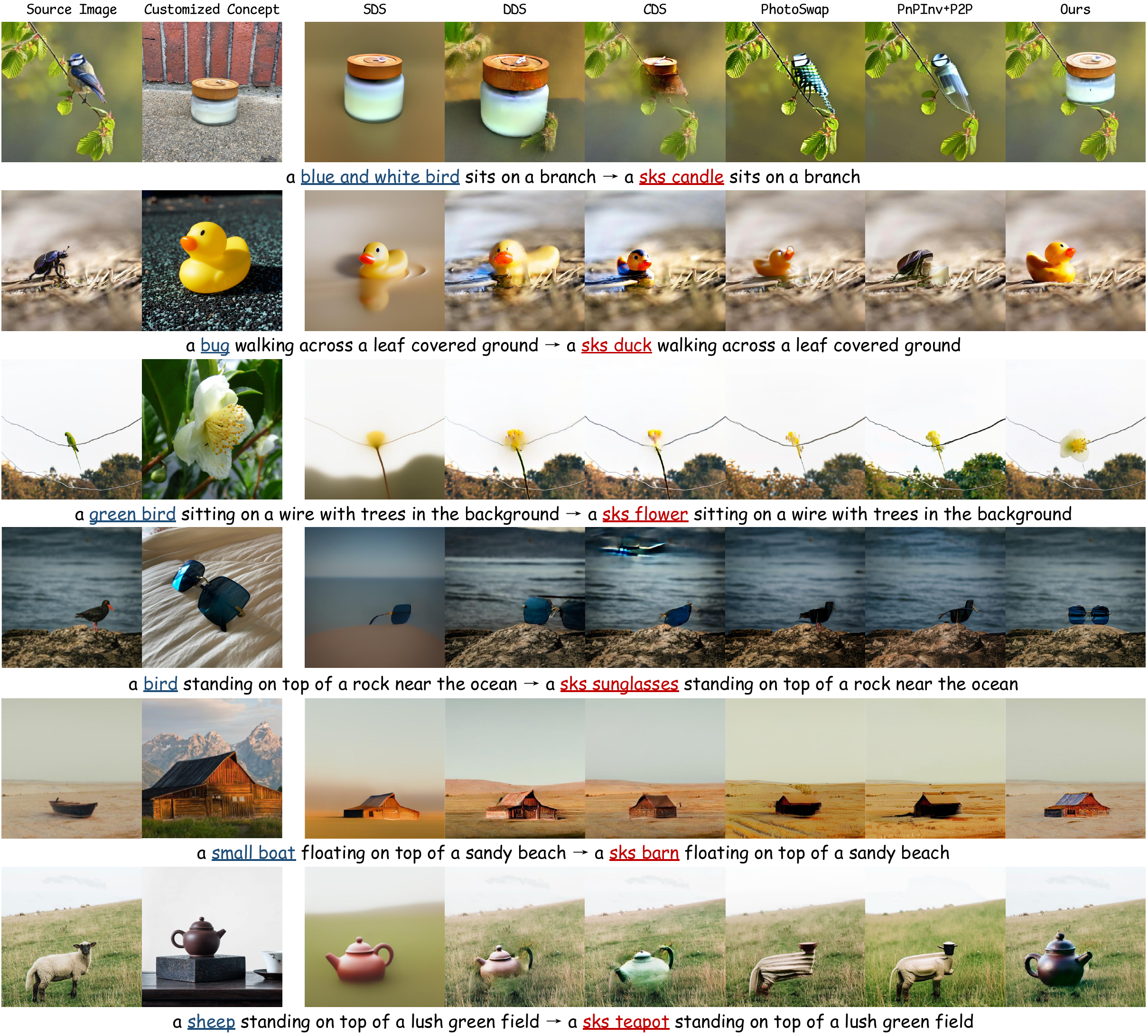}
  \caption{Qualitative comparisons between our \ours and other methods. 
  More qualitative results as well as the used bboxes can be found in 
\textit{Supp.}}
  \vspace{-0.6cm}
  \label{fig:quality}
\end{figure*}

\begin{table*}[th]
\centering
\tabcolsep=2mm
\caption{Quantitative comparisons. 
Our method outperforms all the compared methods in all the selected metrics. 
\textcolor{Red}{\textbf{Red}} stands for the best result, \textcolor{Blue}{\textbf{Blue}} stands for the second best result.}
\fontsize{8pt}{8pt}\selectfont
\resizebox{\linewidth}{!}{
\begin{tabular}{l c|  c  c c c| c c}
    \toprule
     & \multicolumn{1}{c}{\textbf{FG}} & \multicolumn{4}{c}{\textbf{BG}}& \multicolumn{2}{c}{\textbf{Overall}}\\
     \cmidrule(lr){2-2} \cmidrule(lr){3-6} \cmidrule(lr){7-8}
      \multirow{-2}{*}{\textbf{Method}}  & \rotatebox{0}{\textbf{CLIP-I} $\uparrow$}  & \rotatebox{0}{\textbf{PSNR} $\uparrow$}  &\rotatebox{0}{\textbf{LPIPS$_{\times 10^3}$} $\downarrow$}  &\rotatebox{0}{\textbf{MSE} $_{\times 10^4}$ $\downarrow$}  &\rotatebox{0}{\textbf{SSIM} $_{\times 10^2}$ $\uparrow$} & \rotatebox{0}{\textbf{CLIP-T} $\uparrow$}  & \rotatebox{0}{\textbf{Time} (s) $\downarrow$}  \\
     \midrule \midrule
SDS             & \textcolor{Blue}{\textbf{73.70}} & 20.79 & 339.51 & 107.53 & 72.59& 23.53 & 40.37 \\
DDS             & 71.05 & 24.07 & \textcolor{Blue}{\textbf{89.80}}  & 53.08  & \textcolor{Blue}{\textbf{83.44}}& 23.99 & 66.89 \\
CDS             & 71.69 & 23.36 & 90.35  & 63.41  & 83.21& 24.17 & 140.26 \\
PhotoSwap       & 70.15 & 24.24 & 120.62 & 56.64  & 80.56& 22.38 &140.34 \\
PnPInv+P2P    & 70.74 & \textcolor{Blue}{\textbf{24.63}} & 108.22 & \textcolor{Blue}{\textbf{47.49}}  & 82.07& \textcolor{Blue}{\textbf{24.25}} & \textcolor{Blue}{\textbf{37.02}}  \\ \midrule
Ours   & \textcolor{Red}{\textbf{75.00}} & \textcolor{Red}{\textbf{27.39}} & \textcolor{Red}{\textbf{47.68}}  & \textcolor{Red}{\textbf{27.87}}  & \textcolor{Red}{\textbf{86.58}}& \textcolor{Red}{\textbf{25.74}} & \textcolor{Red}{\textbf{19.83}}\\
 \bottomrule
 \end{tabular}}
 \label{tab:Quantitative}
 \end{table*}

\subsubsection{Step-skipping Gradient Updating}\label{sec:SSGU}

After addressing the problem of inconsistency, we turn our attention to the challenge of inefficiency.
As illustrated in \cref{fig:SSGU}, previous methods calculate the gradient at each timestep. 
However, the success of DDIM~\citep{ddim} in accelerating DDPM~\citep{ddpm} motivates us to consider: 
\textit{Can we skip the calculation of gradients at certain timesteps?}
During concept swapping, we observe that the effect of gradient updates on the target image is similar across adjacent timesteps (see detailed results in \textit{Supp.}).
Based on this observation, we propose the step-skipping Gradient Updating (SSGU) strategy.
The key insight of SSGU is that \textit{skipping some gradient calculations does not significantly sacrifice the swapping consistency while considerably improving efficiency.}
As a result, our SSGU calculates gradients at interval timesteps and reuses the previously calculated gradients during the intervening timesteps.

We define our entire pipeline as $\mathcal{F}$, given the source image $x_s$, the timestep $t$, and the intermediate target image $\tilde{x}_t$ at timestep $t$.
We can obtain the gradient $g_t$ and the output intermediate target image $\tilde{x}_{t+1}$ at timestep $t$ as follows:
\begin{gather}
    g_t=\mathcal{F}(x_s,\tilde{x}_t,t),\\
    \tilde{x}_{t+1}=\tilde{x}_t-\eta g_t,
\end{gather}
where $\eta$ is the learning rate.
Our SSGU periodically retains some anchor gradients and skips the forward passes between two anchor gradients. 
The step-skipping period is controlled by the SSGU factor $\lambda$. The set of anchor gradients can be defined as:
\begin{equation}
    \mathbb{G}=\{g_{\lambda k} \},k=0,1\cdots,\lfloor T/\lambda \rfloor, 
\end{equation}
where $T$ is the ended timestep. 
For any intervening timestep $t$, we use its nearest former anchor gradient to update the intermediate target image $\tilde{x}_t$.
Taking $\lambda=2$ as an example, we assume that $t$ is an even number and $g_{t-2}$, $g_t \in \mathbb{G}$.
SSGU updates $\tilde{x}_{t-2}$ and $\tilde{x}_{t-1}$ with the anchor gradient $g_{t-2}$:
\begin{gather}
    g_{t-2}=\mathcal{F}(x_s,\tilde{x}_{t-2},t-2),\\
    \tilde{x}_{t-1}=\tilde{x}_{t-2}-\eta g_{t-2},\\
    \tilde{x}_{t}=\tilde{x}_{t-1}-\eta g_{t-2}.
\end{gather}
For the next timestep $t$, another anchor gradient $g_t$ is used to update $x_t$.
As a result, our SSGU reduces the number of forward passes during the entire concept swapping process to $1 / \lambda$ of the original count. 
Since the forward pass accounts for approximately 95\% of the total inference time (see detailed analysis in \textit{Supp.}), our SSGU can improve the overall inference speed of our method by approximately $\lambda$ times, with minimal effect on swapping consistency (see \cref{abltion:SSGU}). Furthermore, our SSGU can be transferred to other score distillation based methods to improve their efficiency in the same way, which is further discussed in \cref{extension:SSGU}.

\begin{figure*}[tb]
  \centering
  \includegraphics[width=\linewidth]{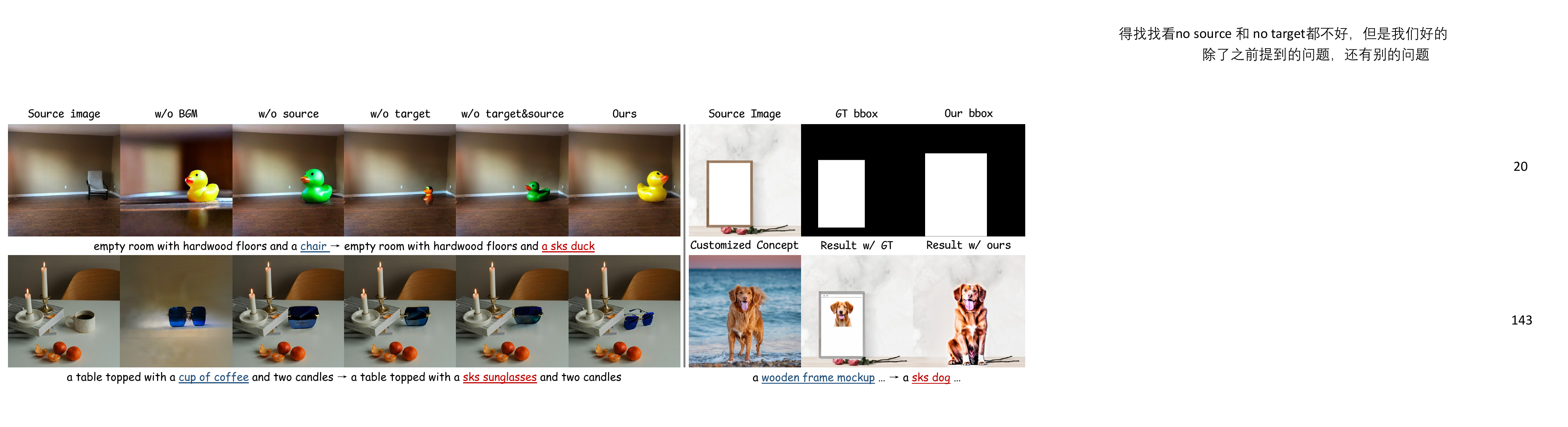}
  \caption{Qualitative results of the ablation study on: \textit{Left}: BGM and SECR. \textit{Right}: different bboxes.}
  \label{fig:cross attn ablatoin}
\end{figure*}

\section{Experiments}
\subsection{Implementation Details}

We conduct the experiments with Stable Diffusion~\citep{SD} v2.1-base on a single RTX3090. 
We use the customized checkpoint from DreamBooth~\citep{db} to introduce concepts.
We set the SSGU factor $\lambda$ to 5, $\alpha$ to 2, $\beta$ to 0.5 and the guidance scale to 7.5.
The bbox is obtained through the first three steps.
Subsequently, we use SGD~\citep{sgd} with a learning rate of 0.1 to optimize for 550 steps of iterations.

\subsection{ConSwapBench}

Despite the significant application potential of customized concept swapping, there is currently no dedicated evaluation benchmark. To meet the needs of comprehensive evaluation, we introduce \csb, the first benchmark dataset specifically designed for customized concept swapping. \csb consists of two sub-benchmarks: ConceptBench and SwapBench. 
ConceptBench comprises 62 images covering 10 different target concepts used for customization, while SwapBench includes 160 real images containing one or more objects to be swapped, serving as source images.
For each image in SwapBench, we use Grounding SAM~\citep{groundsam} to acquire the bbox of the foreground concepts as the ground truth for evaluation purposes.
We apply each customized concept from ConceptBench to perform concept swaps on each image in SwapBench, ultimately generating a total of 1,600 images for evaluation.
More details can be found in \textit{Supp.}

\subsection{Qualitative Comparison}

Since customized concept swapping is a relatively novel task, there are limited methods available for direct comparison. 
Consequently, we include SOTA image editing methods and adapt them for customized concept swapping. 
We include the following methods: (1) \textit{Score distillation based} methods: SDS~\citep{sds}, DDS~\citep{dds}, and CDS~\citep{cds}; (2) \textit{Attention-based} methods: PhotoSwap~\citep{photoswap}, PnPInv~\citep{pnpinversion}, and P2P~\citep{p2p}. We excluded SwapAnything~\citep{swapanything} as it is not publicly available.
The qualitative results are illustrated in \cref{fig:quality}. 
We find that score distillation based methods can accommodate shape variations during concept swapping. 
However, they exhibit poor foreground fidelity (3rd and 6th rows) and lead to unnecessary modifications on the background (1st and 2nd rows). Attention-based methods are unable to manage shape variations (4th row) and also struggle with maintaining background consistency (5th row). 
In contrast, our method demonstrates superior performance in addressing shape variations and maintaining swapping consistency.

\begin{figure*}[tb]
  \centering
  \includegraphics[width=\linewidth]{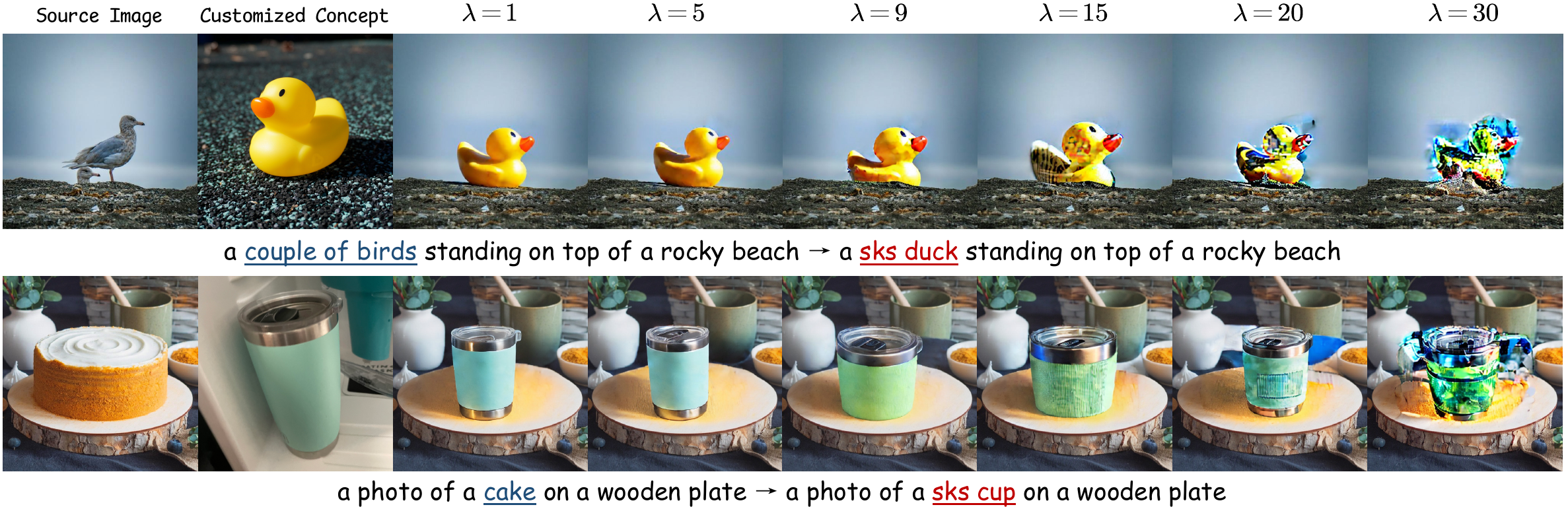}
  \caption{Qualitative results of different SSGU factor $\lambda$. Excessively high $\lambda$ can lead to a decline in foreground consistency.}
  \label{fig:SSGU ablatoin}
\end{figure*}

\subsection{Quantitative Comparison} \label{sec:quantitative}

 We also conduct a thorough quantitative comparison on \csb.
For each generated image, we first use the ground truth bbox in SwapBench to obtain their foreground and background respectively. 
We use seven different metrics to evaluate the methods from three aspects: 
(1) Foreground consistency: We calculate the CLIP Image Score~\citep{clip} between the foreground of generated images and the images of customized concepts.
(2) Background consistency: We use the four metrics, PSNR, LPIPS~\citep{lpips}, MSE, SSIM~\citep{ssim} to evaluate the background consistency.
(3) Overall consistency and efficiency: We calculate the CLIP Text Score~\citep{clipscore} between generated images and target prompts to evaluate the overall prompt consistency. We also report the inference time of each method to evaluate their efficiency.
As shown in \cref{tab:Quantitative}, our method outperforms other methods on all seven metrics.

\subsection{Ablation Study} 

\textbf{BGM.}
To verify the effectiveness of BGM in background preservation, we conduct an ablation study by removing BGM. 
As illustrated in the second column of \cref{fig:cross attn ablatoin}, while our method can still achieve concept swapping without BGM, it causes serious modifications on the background. 
In contrast, our full method not only maintains high foreground fidelity but also effectively preserves the background consistency. 
We further conduct a quantitative analysis of the background consistency and prompt consistency, as shown in \cref{tab:BGM ablation}. Our full method outperforms in all metrics.
\label{sec:ablation bbox}

\begin{table}[h]
\centering
  \tabcolsep=2pt
  \caption{Quantitative ablation results of BGM.} 
  \fontsize{8pt}{8pt}\selectfont
\resizebox{\linewidth}{!}
{
    \begin{tabular}{l|c c c c c}
\toprule
    \textbf{Method} & \rotatebox{0}{\textbf{PSNR} $\uparrow$}  &\rotatebox{0}{\textbf{LPIPS} $\downarrow$}  &\rotatebox{0}{\textbf{MSE}  $\downarrow$}  &\rotatebox{0}{\textbf{SSIM}  $\uparrow$} & \textbf{CLIP-T} $\uparrow$  \\
    \midrule \midrule
    w/o BGM & 18.03 & 249.24 & 184.19 & 72.25 & 23.08   \\ 
    Ours &\textbf{27.39} & \textbf{47.68}  & \textbf{27.87}  & \textbf{86.58}& \textbf{25.74}   \\
\bottomrule
    \end{tabular}
    }
  \label{tab:BGM ablation}
    
\end{table}

\noindent\textbf{Automatic bounding box detection mechanism.}
We further verify the effectiveness of our boxes by using ground truth (GT) bboxes from SwapBench to replace the automatically obtained bboxes
As shown in \cref{fig:cross attn ablatoin}, although GT bboxes accurately indicate the location of the source concept, they prevent our method from fully swapping the source concept. 
Compared to GT bboxes, our bboxes are relatively larger and can fully cover the source concept, thereby facilitating complete concept swapping. 
We also provide quantitative comparisons of different bboxes.
As shown in \cref{tab:bbox abltion}, Gen stands for the generation bboxes, while Eva stands for the evaluation bboxes.
The two types of bboxes do not significantly affect background preservation in our method, whereas our bboxes perform better than GT bboxes on the foreground metric. 
More detailed comparisons can be found in \textit{Supp.}.

\begin{table}[th]
\centering
  \tabcolsep=2pt
  \caption{Quantitative ablation results of different bboxes.} 
  \fontsize{8pt}{8pt}\selectfont
\resizebox{\linewidth}{!}
{
    \begin{tabular}{c c | c c c c c c}
\toprule
     \rotatebox{0}{\textbf{Gen}}&\rotatebox{0}{\textbf{Eva}}& \rotatebox{0}{\textbf{CLIP-I} $\uparrow$}  & \rotatebox{0}{\textbf{PSNR} $\uparrow$}  &\rotatebox{0}{\textbf{LPIPS} $\downarrow$}  &\rotatebox{0}{\textbf{MSE}  $\downarrow$}  &\rotatebox{0}{\textbf{SSIM}  $\uparrow$} & \textbf{CLIP-T} $\uparrow$  \\
    \midrule \midrule
    Ours& GT & 75.00 & 27.39 & 47.68 & 27.87 & 86.58 & \textbf{25.74}    \\ 
    GT & GT & 75.79 & 31.46 & 32.61 & 13.51 & 87.62 & 25.53    \\ 
    Ours & Ours &\textbf{77.72} & \textbf{31.64} & \textbf{31.17} & \textbf{13.28} & \textbf{88.21}& \textbf{25.74}    \\
\bottomrule
    \end{tabular}
    }
    \label{tab:bbox abltion}
    
\end{table}

\begin{figure*}[htb]
  \centering
  \includegraphics[width=\linewidth]{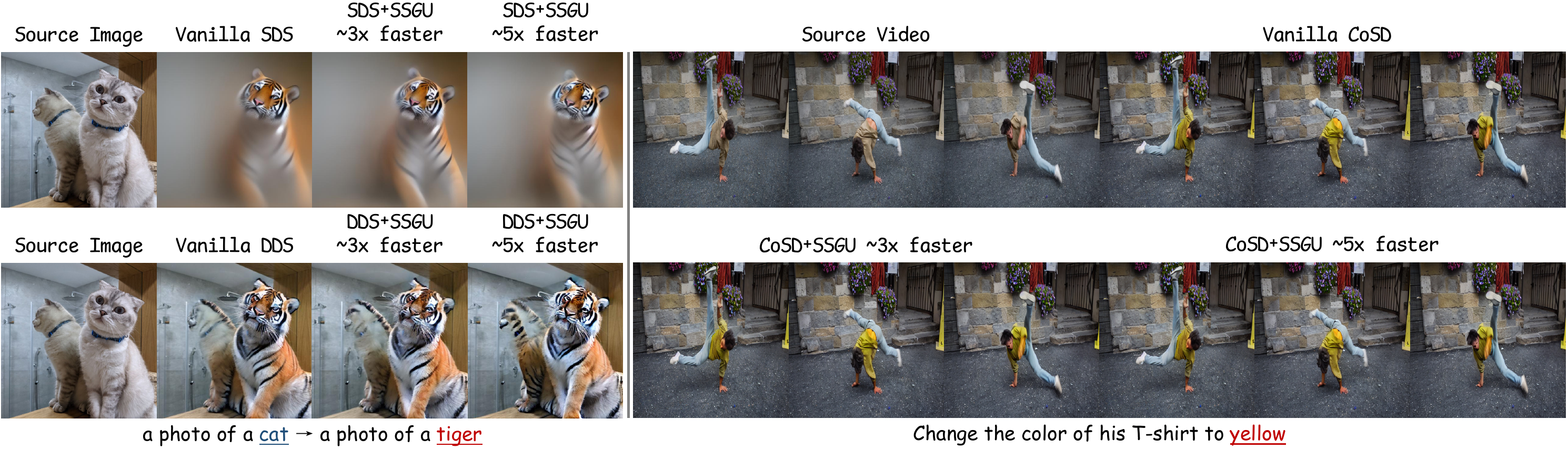}
  \caption{Qualitative results of extending our SSGU to other score distillation based methods.
  }
  \label{fig:SSGU Extension}
\end{figure*}

\begin{figure*}[tb]
  \centering
  \includegraphics[width=\linewidth]{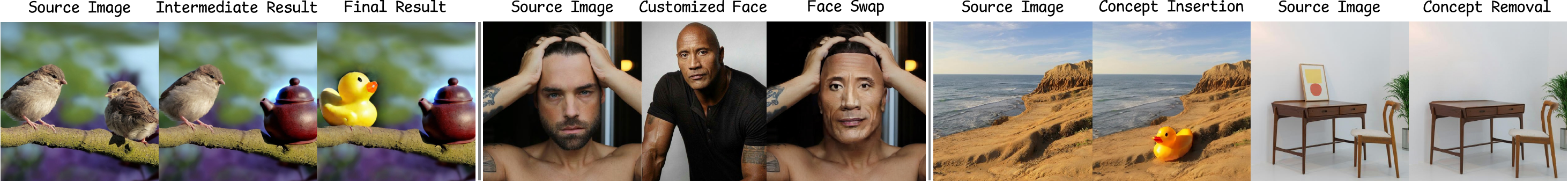}
  \caption{\ours can be extended to other tasks such as: \textit{Left}: Multi-Concept Swapping. \textit{Middle}: Human Face Swapping. \textit{Right}: Concept Insertion and Removal.}
  \label{fig:three extension}
\end{figure*}

\begin{figure}[h]
\centering
\includegraphics[width=\linewidth]{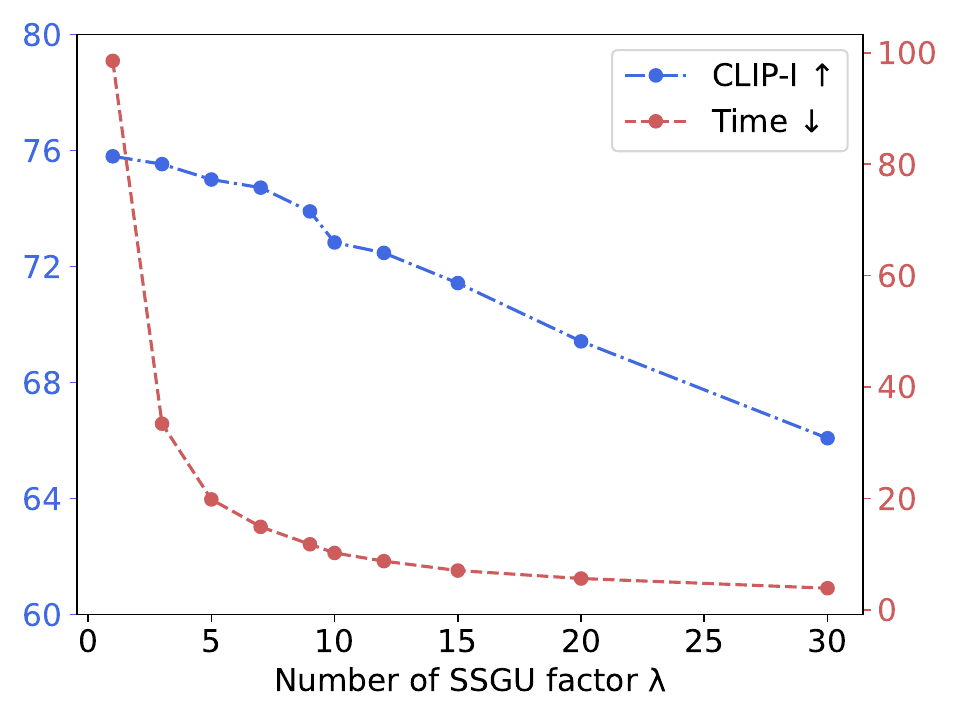}
\caption{Quantitative results on the ablation study of the SSGU factor $\lambda$. 
  }

  \label{fig:abltion on interval}
\end{figure}

\noindent\textbf{SECR.}
To verify the effectiveness of SECR, we conduct ablation studies including removing SECR from (1) source branch (w/o source), (2) target branch (w/o target), (3) both (w/o source \& target).
The visualization results are illustrated in columns 3 to 5 of \cref{fig:cross attn ablatoin}. Although all methods preserve the background well, they show reduced foreground fidelity.
Additionally, we perform a quantitative analysis of their foreground consistency and prompt consistency. 
The results presented in \cref{tab:SECR abltion} indicate that our full method exhibits superior performance.

\begin{table}[h]
\centering
  \tabcolsep=2pt
  \caption{Quantitative ablation results of SECR.} 
  \fontsize{8pt}{8pt}\selectfont
{
    \begin{tabular}{l|c c c }
\toprule
    \textbf{Method} & \textbf{CLIP-I} $\uparrow$ & \textbf{CLIP-T} $\uparrow$  \\
    \midrule \midrule
    w/o source & 73.70 & 25.47  \\
    w/o target & 73.40 & 25.52  \\
    w/o source\&target & 72.42 & 25.21  \\ \midrule
    Ours & \textbf{75.00} & \textbf{25.74}  \\
\bottomrule
    \end{tabular}
    }
  \label{tab:SECR abltion}
    
\end{table}

\noindent\textbf{SSGU.}
\label{abltion:SSGU}
To verify the effectiveness of our proposed SSGU, we first visualize the images generated under different SSGU factors. As shown in \cref{fig:SSGU ablatoin}, $\lambda=1$ indicates that SSGU is not used. When $\lambda \leq 9$, the SSGU can preserve foreground and background consistency well while improving the efficiency of our method.
As $\lambda$ increases, the images exhibit more artifacts due to excessive neglect of gradients. Therefore, identifying an optimal SSGU factor $\lambda$ is crucial. We further conduct a detailed quantitative analysis of different $\lambda$ values on foreground consistency and efficiency (see complete results in \textit{Supp.}), as illustrated in \cref{fig:abltion on interval}, where the $x$-axis represents different $\lambda$ values and the $y$-axis represents the respective metric outcomes. 
When SSGU is not used, our method achieves the best swapping consistency but the lowest efficiency.
As $\lambda$ increases, our SSGU sacrifices certain swapping consistency but significantly improves efficiency. 
To balance consistency and efficiency, we ultimately select $\lambda=5$ for our final model.

\subsection{Applications of \ours}

\noindent\textbf{Multi-concept swapping.}
Our \ours can be easily extended to facilitate multi-concept swaps by sequentially performing multiple single-concept swaps. 
As shown in the left of \cref{fig:three extension}, our method can swap each concept within the image with both foreground and background consistency.

\noindent\textbf{Human face swapping.}
\ours demonstrates exceptional capabilities in human face swapping. As shown in the middle of \cref{fig:three extension}, with customized face models from CivitAI~\citep{civitai}, users can seamlessly replace the face in a source image with a customized target face.

\label{sec:concept insert remove}
\noindent\textbf{Concept insertion and removal.} 
In addition to concept swapping, our method also supports concept insertion and removal. 
For concept insertion, we employ the same procedure as concept swapping. For concept removal, we adjust the target prompt and the target semantic input $p_t$ of SECR to a null prompt.
The results in the right of \cref{fig:three extension} further demonstrate the versatility of our method.

\begin{table}[h]
\centering
  \tabcolsep=2pt
  \caption{Quantitative results of SSGU extension.} 
  \fontsize{8pt}{8pt}\selectfont
{
    \begin{tabular}{l|c c c }
\toprule
    \textbf{Method} & \textbf{w/o SSGU} & \textbf{w/ SSGU $\lambda=3$} & \textbf{w/ SSGU $\lambda=5$} \\
    \midrule \midrule
    SDS & 40.37s & 14.12s & 8.62s \\
    DDS & 66.89s & 22.65s & 13.90s \\
    CoSD & 344.76s & 128.97s & 79.15s \\
\bottomrule
    \end{tabular}
    }
    \label{tab:SSGU extension}
\end{table}

\label{extension:SSGU}
\noindent\textbf{Accelerating other methods.}
SSGU can be transferred to other score distillation based methods to enhance their efficiency. 
We select three representative methods: SDS~\citep{sds}, DDS~\citep{dds} for image editing, and CoSD~\cite{collaborative} for video editing. 
As shown in \cref{fig:SSGU Extension}, combining these methods with SSGU can significantly improve their efficiency while almost not altering the generation quality.
We further conduct a quantitative analysis to assess the transferability of the proposed SSGU, as presented in \cref{tab:SSGU extension}.


\section{Conclusion}
This paper introduces \ours, a novel framework for precise and efficient customized concept swap. Our BGM and SECR collaborate to maintain both background and foreground consistency.
Furthermore, we propose the SSGU to eliminate redundant computation and improve efficiency. Finally, we introduce \csb, a comprehensive benchmark dataset for customized concept swapping. The impressive performance of \ours demonstrates its effectiveness.
We hope our \ours can inspire future research, particularly in efficiently managing concept swapping with obvious shape variance. Future work could focus on
(1) extending image-based customized concept swapping to the video domain;
(2) enhancing the images of target concepts with low-level methods~\citep{he2023retidiff,fang2024realworld,he2024diffusion}; and
(3) achieving more lightweight and precise concept swapping.
{
    \small
    \bibliographystyle{ieeenat_fullname}
    \bibliography{main}

\begin{thebibliography}{54}
\providecommand{\natexlab}[1]{#1}
\providecommand{\url}[1]{\texttt{#1}}
\expandafter\ifx\csname urlstyle\endcsname\relax
  \providecommand{\doi}[1]{doi: #1}\else
  \providecommand{\doi}{doi: \begingroup \urlstyle{rm}\Url}\fi

\bibitem[Brooks et~al.(2023)Brooks, Holynski, and Efros]{instructpix2pix}
Tim Brooks, Aleksander Holynski, and Alexei~A Efros.
\newblock Instructpix2pix: Learning to follow image editing instructions.
\newblock In \emph{Proceedings of the IEEE/CVF Conference on Computer Vision and Pattern Recognition}, pages 18392--18402, 2023.

\bibitem[Cao et~al.(2023)Cao, Wang, Qi, Shan, Qie, and Zheng]{masactrl}
Mingdeng Cao, Xintao Wang, Zhongang Qi, Ying Shan, Xiaohu Qie, and Yinqiang Zheng.
\newblock Masactrl: Tuning-free mutual self-attention control for consistent image synthesis and editing.
\newblock In \emph{Proceedings of the IEEE/CVF International Conference on Computer Vision}, pages 22560--22570, 2023.

\bibitem[Chang et~al.(2024)Chang, Chang, and Ye]{groundscore}
Hangeol Chang, Jinho Chang, and Jong~Chul Ye.
\newblock Ground-a-score: Scaling up the score distillation for multi-attribute editing.
\newblock \emph{arXiv preprint arXiv:2403.13551}, 2024.

\bibitem[Choi et~al.(2023)Choi, Choi, Kim, Kim, and Yoon]{customedit}
Jooyoung Choi, Yunjey Choi, Yunji Kim, Junho Kim, and Sungroh Yoon.
\newblock Custom-edit: Text-guided image editing with customized diffusion models.
\newblock \emph{arXiv preprint arXiv:2305.15779}, 2023.

\bibitem[{Civitai}(2024)]{civitai}
{Civitai}.
\newblock Civitai.
\newblock \url{https://civitai.com/}, 2024.

\bibitem[Couairon et~al.(2022)Couairon, Verbeek, Schwenk, and Cord]{diffedit}
Guillaume Couairon, Jakob Verbeek, Holger Schwenk, and Matthieu Cord.
\newblock Diffedit: Diffusion-based semantic image editing with mask guidance.
\newblock \emph{arXiv preprint arXiv:2210.11427}, 2022.

\bibitem[Dhariwal and Nichol(2021)]{diffbgan}
Prafulla Dhariwal and Alexander Nichol.
\newblock Diffusion models beat gans on image synthesis.
\newblock \emph{Advances in neural information processing systems}, 34:\penalty0 8780--8794, 2021.

\bibitem[Dong et~al.(2023)Dong, Xue, Duan, and Han]{dong2023prompt}
Wenkai Dong, Song Xue, Xiaoyue Duan, and Shumin Han.
\newblock Prompt tuning inversion for text-driven image editing using diffusion models.
\newblock In \emph{Proceedings of the IEEE/CVF International Conference on Computer Vision}, pages 7430--7440, 2023.

\bibitem[Fang et~al.(2024)Fang, He, Xiao, Zhang, Tang, Zhang, Li, and Li]{fang2024realworld}
Chengyu Fang, Chunming He, Fengyang Xiao, Yulun Zhang, Longxiang Tang, Yuelin Zhang, Kai Li, and Xiu Li.
\newblock Real-world image dehazing with coherence-based pseudo labeling and cooperative unfolding network.
\newblock In \emph{The Thirty-eighth Annual Conference on Neural Information Processing Systems}, 2024.

\bibitem[Gal et~al.(2022)Gal, Alaluf, Atzmon, Patashnik, Bermano, Chechik, and Cohen-Or]{TI}
Rinon Gal, Yuval Alaluf, Yuval Atzmon, Or Patashnik, Amit~H Bermano, Gal Chechik, and Daniel Cohen-Or.
\newblock An image is worth one word: Personalizing text-to-image generation using textual inversion.
\newblock \emph{arXiv preprint arXiv:2208.01618}, 2022.

\bibitem[Geng et~al.(2024)Geng, Yang, Hang, Li, Gu, Zhang, Bao, Zhang, Li, Hu, et~al.]{instructdiffusion}
Zigang Geng, Binxin Yang, Tiankai Hang, Chen Li, Shuyang Gu, Ting Zhang, Jianmin Bao, Zheng Zhang, Houqiang Li, Han Hu, et~al.
\newblock Instructdiffusion: A generalist modeling interface for vision tasks.
\newblock In \emph{Proceedings of the IEEE/CVF Conference on Computer Vision and Pattern Recognition}, pages 12709--12720, 2024.

\bibitem[Goodfellow et~al.(2020)Goodfellow, Pouget-Abadie, Mirza, Xu, Warde-Farley, Ozair, Courville, and Bengio]{gan}
Ian Goodfellow, Jean Pouget-Abadie, Mehdi Mirza, Bing Xu, David Warde-Farley, Sherjil Ozair, Aaron Courville, and Yoshua Bengio.
\newblock Generative adversarial networks.
\newblock \emph{Communications of the ACM}, 63\penalty0 (11):\penalty0 139--144, 2020.

\bibitem[Gu et~al.(2024{\natexlab{a}})Gu, Wang, Zhao, Fu, Xiong, Liu, Zhang, Zhang, Zhang, Jung, et~al.]{photoswap}
Jing Gu, Yilin Wang, Nanxuan Zhao, Tsu-Jui Fu, Wei Xiong, Qing Liu, Zhifei Zhang, He Zhang, Jianming Zhang, HyunJoon Jung, et~al.
\newblock Photoswap: Personalized subject swapping in images.
\newblock \emph{Advances in Neural Information Processing Systems}, 36, 2024{\natexlab{a}}.

\bibitem[Gu et~al.(2024{\natexlab{b}})Gu, Wang, Zhao, Xiong, Liu, Zhang, Zhang, Zhang, Jung, and Wang]{swapanything}
Jing Gu, Yilin Wang, Nanxuan Zhao, Wei Xiong, Qing Liu, Zhifei Zhang, He Zhang, Jianming Zhang, HyunJoon Jung, and Xin~Eric Wang.
\newblock Swapanything: Enabling arbitrary object swapping in personalized visual editing.
\newblock \emph{arXiv preprint arXiv:2404.05717}, 2024{\natexlab{b}}.

\bibitem[Guo and Lin(2024)]{guo2024focus}
Qin Guo and Tianwei Lin.
\newblock Focus on your instruction: Fine-grained and multi-instruction image editing by attention modulation.
\newblock In \emph{Proceedings of the IEEE/CVF Conference on Computer Vision and Pattern Recognition}, pages 6986--6996, 2024.

\bibitem[He et~al.(2023)He, Fang, Zhang, Li, Tang, You, Xiao, Guo, and Li]{he2023retidiff}
Chunming He, Chengyu Fang, Yulun Zhang, Kai Li, Longxiang Tang, Chenyu You, Fengyang Xiao, Zhenhua Guo, and Xiu Li.
\newblock Reti-diff: Illumination degradation image restoration with retinex-based latent diffusion model.
\newblock 2023.

\bibitem[He et~al.(2024)He, Shen, Fang, Xiao, Tang, Zhang, Zuo, Guo, and Li]{he2024diffusion}
Chunming He, Yuqi Shen, Chengyu Fang, Fengyang Xiao, Longxiang Tang, Yulun Zhang, Wangmeng Zuo, Zhenhua Guo, and Xiu Li.
\newblock Diffusion models in low-level vision: A survey.
\newblock \emph{arXiv preprint arXiv:2406.11138}, 2024.

\bibitem[Hertz et~al.(2022)Hertz, Mokady, Tenenbaum, Aberman, Pritch, and Cohen-Or]{p2p}
Amir Hertz, Ron Mokady, Jay Tenenbaum, Kfir Aberman, Yael Pritch, and Daniel Cohen-Or.
\newblock Prompt-to-prompt image editing with cross attention control.
\newblock \emph{arXiv preprint arXiv:2208.01626}, 2022.

\bibitem[Hertz et~al.(2023)Hertz, Aberman, and Cohen-Or]{dds}
Amir Hertz, Kfir Aberman, and Daniel Cohen-Or.
\newblock Delta denoising score.
\newblock In \emph{Proceedings of the IEEE/CVF International Conference on Computer Vision}, pages 2328--2337, 2023.

\bibitem[Hessel et~al.(2021)Hessel, Holtzman, Forbes, Bras, and Choi]{clipscore}
Jack Hessel, Ari Holtzman, Maxwell Forbes, Ronan~Le Bras, and Yejin Choi.
\newblock Clipscore: A reference-free evaluation metric for image captioning.
\newblock \emph{arXiv preprint arXiv:2104.08718}, 2021.

\bibitem[Ho et~al.(2020)Ho, Jain, and Abbeel]{ddpm}
Jonathan Ho, Ajay Jain, and Pieter Abbeel.
\newblock Denoising diffusion probabilistic models.
\newblock \emph{Advances in neural information processing systems}, 33:\penalty0 6840--6851, 2020.

\bibitem[Huang et~al.(2023)Huang, Tu, and Xu]{huang2023pfb}
Wenjing Huang, Shikui Tu, and Lei Xu.
\newblock Pfb-diff: Progressive feature blending diffusion for text-driven image editing.
\newblock \emph{arXiv preprint arXiv:2306.16894}, 2023.

\bibitem[Huang et~al.(2024)Huang, Xie, Wang, Yuan, Cun, Ge, Zhou, Dong, Huang, Zhang, et~al.]{smartedit}
Yuzhou Huang, Liangbin Xie, Xintao Wang, Ziyang Yuan, Xiaodong Cun, Yixiao Ge, Jiantao Zhou, Chao Dong, Rui Huang, Ruimao Zhang, et~al.
\newblock Smartedit: Exploring complex instruction-based image editing with multimodal large language models.
\newblock In \emph{Proceedings of the IEEE/CVF Conference on Computer Vision and Pattern Recognition}, pages 8362--8371, 2024.

\bibitem[Ju et~al.(2024)Ju, Zeng, Bian, Liu, and Xu]{pnpinversion}
Xuan Ju, Ailing Zeng, Yuxuan Bian, Shaoteng Liu, and Qiang Xu.
\newblock Pnp inversion: Boosting diffusion-based editing with 3 lines of code.
\newblock In \emph{The Twelfth International Conference on Learning Representations}, 2024.

\bibitem[Kim et~al.(2023)Kim, Lee, Choi, Jeong, Sohn, and Shin]{collaborative}
Subin Kim, Kyungmin Lee, June~Suk Choi, Jongheon Jeong, Kihyuk Sohn, and Jinwoo Shin.
\newblock Collaborative score distillation for consistent visual synthesis.
\newblock \emph{arXiv preprint arXiv:2307.04787}, 2023.

\bibitem[Kumari et~al.(2023)Kumari, Zhang, Zhang, Shechtman, and Zhu]{cd}
Nupur Kumari, Bingliang Zhang, Richard Zhang, Eli Shechtman, and Jun-Yan Zhu.
\newblock Multi-concept customization of text-to-image diffusion.
\newblock In \emph{Proceedings of the IEEE/CVF Conference on Computer Vision and Pattern Recognition}, pages 1931--1941, 2023.

\bibitem[Li et~al.(2024)Li, Nie, Chen, Jiang, Wu, Lin, Liu, Peng, Wang, and Zheng]{tuning}
Pengzhi Li, Qiang Nie, Ying Chen, Xi Jiang, Kai Wu, Yuhuan Lin, Yong Liu, Jinlong Peng, Chengjie Wang, and Feng Zheng.
\newblock Tuning-free image customization with image and text guidance.
\newblock \emph{arXiv preprint arXiv:2403.12658}, 2024.

\bibitem[Li et~al.(2023)Li, Ku, Wei, and Chen]{dreamedit}
Tianle Li, Max Ku, Cong Wei, and Wenhu Chen.
\newblock Dreamedit: Subject-driven image editing.
\newblock \emph{arXiv preprint arXiv:2306.12624}, 2023.

\bibitem[Liu et~al.(2024)Liu, Wang, Cao, Jia, and Huang]{towards}
Bingyan Liu, Chengyu Wang, Tingfeng Cao, Kui Jia, and Jun Huang.
\newblock Towards understanding cross and self-attention in stable diffusion for text-guided image editing.
\newblock In \emph{Proceedings of the IEEE/CVF Conference on Computer Vision and Pattern Recognition}, pages 7817--7826, 2024.

\bibitem[Ma et~al.(2023)Ma, Cun, He, Qi, Wang, Shan, Li, and Chen]{ma2023magicstick}
Yue Ma, Xiaodong Cun, Yingqing He, Chenyang Qi, Xintao Wang, Ying Shan, Xiu Li, and Qifeng Chen.
\newblock Magicstick: Controllable video editing via control handle transformations.
\newblock \emph{arXiv preprint arXiv:2312.03047}, 2023.

\bibitem[Ma et~al.(2024{\natexlab{a}})Ma, He, Cun, Wang, Chen, Li, and Chen]{ma2024followyourpose}
Yue Ma, Yingqing He, Xiaodong Cun, Xintao Wang, Siran Chen, Xiu Li, and Qifeng Chen.
\newblock Follow your pose: Pose-guided text-to-video generation using pose-free videos.
\newblock In \emph{Proceedings of the AAAI Conference on Artificial Intelligence}, pages 4117--4125, 2024{\natexlab{a}}.

\bibitem[Ma et~al.(2024{\natexlab{b}})Ma, He, Wang, Wang, Qi, Cai, Li, Li, Shum, Liu, et~al.]{ma2024followyourclick}
Yue Ma, Yingqing He, Hongfa Wang, Andong Wang, Chenyang Qi, Chengfei Cai, Xiu Li, Zhifeng Li, Heung-Yeung Shum, Wei Liu, et~al.
\newblock Follow-your-click: Open-domain regional image animation via short prompts.
\newblock \emph{arXiv preprint arXiv:2403.08268}, 2024{\natexlab{b}}.

\bibitem[Ma et~al.(2024{\natexlab{c}})Ma, Liu, Wang, Pan, He, Yuan, Zeng, Cai, Shum, Liu, et~al.]{ma2024followyouremoji}
Yue Ma, Hongyu Liu, Hongfa Wang, Heng Pan, Yingqing He, Junkun Yuan, Ailing Zeng, Chengfei Cai, Heung-Yeung Shum, Wei Liu, et~al.
\newblock Follow-your-emoji: Fine-controllable and expressive freestyle portrait animation.
\newblock \emph{arXiv preprint arXiv:2406.01900}, 2024{\natexlab{c}}.

\bibitem[Miyake et~al.(2023)Miyake, Iohara, Saito, and Tanaka]{negative}
Daiki Miyake, Akihiro Iohara, Yu Saito, and Toshiyuki Tanaka.
\newblock Negative-prompt inversion: Fast image inversion for editing with text-guided diffusion models.
\newblock \emph{arXiv preprint arXiv:2305.16807}, 2023.

\bibitem[Mokady et~al.(2023)Mokady, Hertz, Aberman, Pritch, and Cohen-Or]{null}
Ron Mokady, Amir Hertz, Kfir Aberman, Yael Pritch, and Daniel Cohen-Or.
\newblock Null-text inversion for editing real images using guided diffusion models.
\newblock In \emph{Proceedings of the IEEE/CVF Conference on Computer Vision and Pattern Recognition}, pages 6038--6047, 2023.

\bibitem[Nam et~al.(2024)Nam, Kwon, Park, and Ye]{cds}
Hyelin Nam, Gihyun Kwon, Geon~Yeong Park, and Jong~Chul Ye.
\newblock Contrastive denoising score for text-guided latent diffusion image editing.
\newblock In \emph{Proceedings of the IEEE/CVF Conference on Computer Vision and Pattern Recognition}, pages 9192--9201, 2024.

\bibitem[Nguyen et~al.(2024)Nguyen, Vu, Tran, and Nguyen]{datasetdiff}
Quang Nguyen, Truong Vu, Anh Tran, and Khoi Nguyen.
\newblock Dataset diffusion: Diffusion-based synthetic data generation for pixel-level semantic segmentation.
\newblock \emph{Advances in Neural Information Processing Systems}, 36, 2024.

\bibitem[Nichol et~al.(2021)Nichol, Dhariwal, Ramesh, Shyam, Mishkin, McGrew, Sutskever, and Chen]{glide}
Alex Nichol, Prafulla Dhariwal, Aditya Ramesh, Pranav Shyam, Pamela Mishkin, Bob McGrew, Ilya Sutskever, and Mark Chen.
\newblock Glide: Towards photorealistic image generation and editing with text-guided diffusion models.
\newblock \emph{arXiv preprint arXiv:2112.10741}, 2021.

\bibitem[Poole et~al.(2022)Poole, Jain, Barron, and Mildenhall]{sds}
Ben Poole, Ajay Jain, Jonathan~T Barron, and Ben Mildenhall.
\newblock Dreamfusion: Text-to-3d using 2d diffusion.
\newblock \emph{arXiv preprint arXiv:2209.14988}, 2022.

\bibitem[Radford et~al.(2021)Radford, Kim, Hallacy, Ramesh, Goh, Agarwal, Sastry, Askell, Mishkin, Clark, et~al.]{clip}
Alec Radford, Jong~Wook Kim, Chris Hallacy, Aditya Ramesh, Gabriel Goh, Sandhini Agarwal, Girish Sastry, Amanda Askell, Pamela Mishkin, Jack Clark, et~al.
\newblock Learning transferable visual models from natural language supervision.
\newblock In \emph{International conference on machine learning}, pages 8748--8763. PMLR, 2021.

\bibitem[Ren et~al.(2024)Ren, Liu, Zeng, Lin, Li, Cao, Chen, Huang, Chen, Yan, et~al.]{groundsam}
Tianhe Ren, Shilong Liu, Ailing Zeng, Jing Lin, Kunchang Li, He Cao, Jiayu Chen, Xinyu Huang, Yukang Chen, Feng Yan, et~al.
\newblock Grounded sam: Assembling open-world models for diverse visual tasks.
\newblock \emph{arXiv preprint arXiv:2401.14159}, 2024.

\bibitem[Robbins and Monro(1951)]{sgd}
Herbert Robbins and Sutton Monro.
\newblock A stochastic approximation method.
\newblock \emph{The annals of mathematical statistics}, pages 400--407, 1951.

\bibitem[Rombach et~al.(2022)Rombach, Blattmann, Lorenz, Esser, and Ommer]{SD}
Robin Rombach, Andreas Blattmann, Dominik Lorenz, Patrick Esser, and Bj{\"o}rn Ommer.
\newblock High-resolution image synthesis with latent diffusion models.
\newblock In \emph{Proceedings of the IEEE/CVF conference on computer vision and pattern recognition}, pages 10684--10695, 2022.

\bibitem[Ruiz et~al.(2023)Ruiz, Li, Jampani, Pritch, Rubinstein, and Aberman]{db}
Nataniel Ruiz, Yuanzhen Li, Varun Jampani, Yael Pritch, Michael Rubinstein, and Kfir Aberman.
\newblock Dreambooth: Fine tuning text-to-image diffusion models for subject-driven generation.
\newblock In \emph{Proceedings of the IEEE/CVF conference on computer vision and pattern recognition}, pages 22500--22510, 2023.

\bibitem[Song et~al.(2020)Song, Meng, and Ermon]{ddim}
Jiaming Song, Chenlin Meng, and Stefano Ermon.
\newblock Denoising diffusion implicit models.
\newblock \emph{arXiv preprint arXiv:2010.02502}, 2020.

\bibitem[Tumanyan et~al.(2023)Tumanyan, Geyer, Bagon, and Dekel]{pnp}
Narek Tumanyan, Michal Geyer, Shai Bagon, and Tali Dekel.
\newblock Plug-and-play diffusion features for text-driven image-to-image translation.
\newblock In \emph{Proceedings of the IEEE/CVF Conference on Computer Vision and Pattern Recognition}, pages 1921--1930, 2023.

\bibitem[Wang et~al.(2024)Wang, Ma, Guo, Xiao, Huang, and Li]{wang2024cove}
Jiangshan Wang, Yue Ma, Jiayi Guo, Yicheng Xiao, Gao Huang, and Xiu Li.
\newblock Cove: Unleashing the diffusion feature correspondence for consistent video editing.
\newblock \emph{arXiv preprint arXiv:2406.08850}, 2024.

\bibitem[Wang et~al.(2004)Wang, Bovik, Sheikh, and Simoncelli]{ssim}
Zhou Wang, Alan~C Bovik, Hamid~R Sheikh, and Eero~P Simoncelli.
\newblock Image quality assessment: from error visibility to structural similarity.
\newblock \emph{IEEE transactions on image processing}, 13\penalty0 (4):\penalty0 600--612, 2004.

\bibitem[Yang et~al.(2023)Yang, Gu, Zhang, Zhang, Chen, Sun, Chen, and Wen]{paint}
Binxin Yang, Shuyang Gu, Bo Zhang, Ting Zhang, Xuejin Chen, Xiaoyan Sun, Dong Chen, and Fang Wen.
\newblock Paint by example: Exemplar-based image editing with diffusion models.
\newblock In \emph{Proceedings of the IEEE/CVF Conference on Computer Vision and Pattern Recognition}, pages 18381--18391, 2023.

\bibitem[Zhang et~al.(2024{\natexlab{a}})Zhang, Mo, Chen, Sun, and Su]{magicbrush}
Kai Zhang, Lingbo Mo, Wenhu Chen, Huan Sun, and Yu Su.
\newblock Magicbrush: A manually annotated dataset for instruction-guided image editing.
\newblock \emph{Advances in Neural Information Processing Systems}, 36, 2024{\natexlab{a}}.

\bibitem[Zhang et~al.(2018)Zhang, Isola, Efros, Shechtman, and Wang]{lpips}
Richard Zhang, Phillip Isola, Alexei~A Efros, Eli Shechtman, and Oliver Wang.
\newblock The unreasonable effectiveness of deep features as a perceptual metric.
\newblock In \emph{Proceedings of the IEEE conference on computer vision and pattern recognition}, pages 586--595, 2018.

\bibitem[Zhang et~al.(2023)Zhang, Xiao, and Huang]{forgedit}
Shiwen Zhang, Shuai Xiao, and Weilin Huang.
\newblock Forgedit: Text guided image editing via learning and forgetting.
\newblock \emph{arXiv preprint arXiv:2309.10556}, 2023.

\bibitem[Zhang et~al.(2024{\natexlab{b}})Zhang, Yang, Zhou, and Wang]{attn-calib}
Yanbing Zhang, Mengping Yang, Qin Zhou, and Zhe Wang.
\newblock Attention calibration for disentangled text-to-image personalization.
\newblock In \emph{Proceedings of the IEEE/CVF Conference on Computer Vision and Pattern Recognition}, pages 4764--4774, 2024{\natexlab{b}}.

\bibitem[Zhu et~al.(2024)Zhu, Li, Ma, He, and Xiu]{multibooth}
Chenyang Zhu, Kai Li, Yue Ma, Chunming He, and Li Xiu.
\newblock Multibooth: Towards generating all your concepts in an image from text.
\newblock \emph{arXiv preprint arXiv:2404.14239}, 2024.

\end{thebibliography}
}


\end{document}